# Adaptability and Homeostasis in the Game of Life interacting with the evolved Cellular Automata

Keisuke Suzuki[1] and Takashi Ikegami[2]

[1]Laboratory for Adaptive Intelligence,

RIKEN Brain Science Institute,

2-1 Hirosawa, Wako, Saitama, Japan

+81-48-467-5274

[2]Department of General Systems Sciences,

The Graduate School of Arts and Sciences, The University of Tokyo,

3-8-1 Komaba, Tokyo, 153-8902, Japan

+81-3-5454-6541

The corresponding author: Keisuke Suzuki ksk@brain.riken.jp

**Abstract**

In this paper we study the emergence of homeostasis in a two-layer system of the Game of Life, in which the Game of Life in the first layer couples with another system of cellular automata in the second layer. Homeostasis is defined here as a space-time dynamic that regulates the number of cells in state-1 in the Game of Life layer. A genetic algorithm is used to evolve the rules of the second layer to control the pattern of the Game of Life. We discovered that there are two antagonistic attractors that control the numbers of cells in state-1 in the first layer. The homeostasis sustained by these attractors are compared with the homeostatic dynamics observed in Daisy World.

# Introduction

Living systems require a stable and sustainable structure to survive in unstable and open environments. The maintenance of such a structure is called "homeostasis". This term was coined by Cannon (1932) and Bernard(1957), and has become one of the central themes in cybernetic studies(Wiener, 1948). Several mechanisms underlying homeostasis have been proposed and these have become a guiding principle of our everyday technology. For example, the idea of positive/negative feedback loops came from the cybernetics field.

The controlling mechanisms of homeostasis have been revealed, but very few studies have been done on the self-organization aspect of homeostasis. Exceptions can be found in ecological homeostasis. Many people have studied ecological homeostasis, in particular after Lovelock (1972) proposed his Gaia hypothesis. The Gaia hypothesis posits that the complex and global network of living/nonliving systems self-organizes to achieve homeostasis. The Gaia hypothesis has been theoretically examined by Watson and Lovelock (1983) by developing the Daisy World model, a simple implementation of the Gaia theory. In the Daisy World, temperature is sustained at a certain range independent of the environmental temperature. This is done by complex interaction of two kinds of daises which have different albedos. Harvey (2004) calls the mechanism underlying the Daisy World a "rein control", a controlling mechanism that serves to pull the temperature toward the viability zone.

With respect to the adaptability of homeo-systems, Ashby(1960) proposed an interesting design principle for the brain and for life forms in general that was mainly driven by homeostasis. He posited that the adaptive behavior of life is only an outcome of homeostatic properties, and proposed a different type of homeostatic system called an "ultra-stable system". This new system has two feedback loops. The primary feedback loop is driven by a mutual interaction between an organism's complex sensory and motor channels and the environment. The other feedback loop develops from the interaction between viability constraints and the relevant reacting parts via the essential variables that control the reacting parts. Usually, the second feedback loop is intended to change the meta-parameters of the primary feedback-loop. When parameter values are outside the viability constraints, the second feedback loop adjusts the essential parameters so that the system will move toward a more stable state. This second feedback loop was identified in light of the self-mobility in our previous study, which we name "homeo-dynamics" (Ikegami and Suzuki, 2008). That is, two dynamics co-exist in the same system with different time scales that cooperatively control the homeostasis by maintaining fluctuations in the system. In other words, we need both stable and unstable dynamics to develop homeostasis and adaptation at the same time.

When we regard living systems as natural computing process, such computation may also require concurrent stable and unstable dynamics. Given this, the natural system should possess robustness (e.g. a required degree of insensitivity to environmental changes) in order to achieve certain task, and at the same time, it also needs instability to become sensitive to certain environmental changes. The former environmental change is called "noise" and the latter is called "information". The difference between noise and information is not determined beforehand. It can only be progressively defined.

In this paper, we study the notion of homeostasis and adaptation using Conway's Game of Life. A major drawback of most homeostatic models, including ours, is that many systems test their homeostatic nature in too-stable environment. What this means is that such systems can survive without paying significant costs (or in other words, the system never dies). Therefore, our challenge is to see how homeostasis can emerge even in a very unstable world, such as that of the Game of Life. We intend to see how Daisy World is emerging in the evolutionary process, producing dynamical patterns in a bottom-up way, and not just using the predefined components. We propose a coupled cellular automata (CA) model to study evolution of homeostatic behavior in an unstable bit space.

A basic strategy is to apply another rule set in a site specific way on top of the Life rule. We will show that we can evolve different CA sets for stabilizing (insensitive) and memorizing (sensitive) conditions. Then we will discuss the homeostatic result with respect to the Daisy World setup, where black and white daisies are making the homeostatic adjustment. In the Daisy World setup, the population of black and white daisies and their growth as a simple function of temperature is assumed. In this paper, we are evolving corresponding daisy-like behaviors in the second CA layer. The evolving CA rules and their spatial configuration generate the homeostatic dynamics.

In the next section we describe how to use the Game of Life to study homeostasis. In Section 3, we describe the evolved CA rules and how they work. Finally, in Section 4, we analyze the results. In Section 5, we discuss the observed characteristics of homeostasis and adaptation with respect to the daisy world setup.

## The Model

The basic idea of the model is inspired from work by (Taylor, 2004). In his model, the system under examination consists of two layers of cellular automata: one layer is the Game of Life, and the other layer called 'Genome' serves to control the Game of Life pattern. The entire dynamics is composed of alternative updating of Life rule and the cellular automata in the Genome layer. Taylor evolved the rules of this Genome layer using an evolutionary algorithm to control a virtual sensori-motor flow arranged on the Game of Life. It is defined as an 'input' cell site, and 'target' area consists of several cell sites, where the input cell and output area are apart each other. The goal of evolution in Taylor's work was that when state-1 occurs in an input cell site, the target area should have many state-1 cells. This contingency between an input state and output states is mediated by the intermediate area, in which some of the cell sites are governed by the rules in the genome layer.

The purpose of Taylor's study was to examine an unseparated body-environment boundary and to observe the emergence of the boundary itself. We will also investigate this point, but here we will use his approach to study homeostasis. Our setup is described below.

The model consists of 2D cellular automata (CA) running the Game of Life. Extra rules can override Life states in a certain part of the CA space. Those rules consist of set of condition and output (CA rules) and the coordinates in the space where they are applied (Fig.1). The condition of the CA rules used in this study, is a so-called "totalistic rule", which only takes into account the number of neighboring state-1 cells, and not any specific neighboring patterns. These rules are grouped and encoded in one genome, which is the unit that will undergo evolutionary selection in the genetic algorithm. Figure 1 illustrates the relationship between the rules and the genome. Note that in Taylor's model, the gene consists of condition and output, which have specific state and position of neighboring cells so that the genes activate and update output state only when the specific condition is satisfied, not every time steps. In our model, all the rules are activated and making output 1 or 0 at every time step, because the gene contains all conditions for the states of the cell. In addition, Taylor uses "temporal" gene which activates only when a certain timestep comes. We do not use this temporal rule in our model.

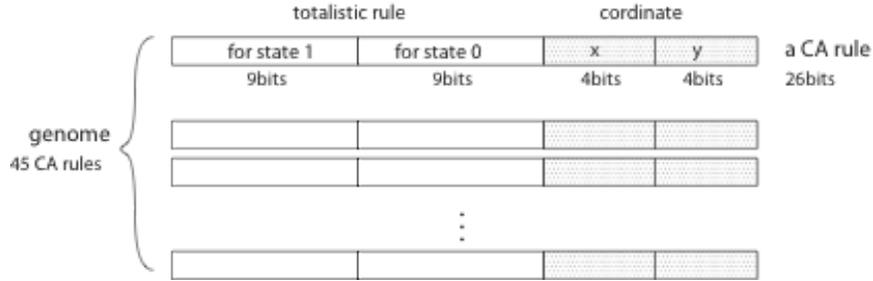

Figure 1: A CA rule consists of a 26-bit binary string, 18 bits of which encode the totalistic CA rules. The remaining bits encode the coordinates of the site where the rule activates. A group of these CA rules is called a genome. The simulation is run using Game of Life dynamics and one genome.

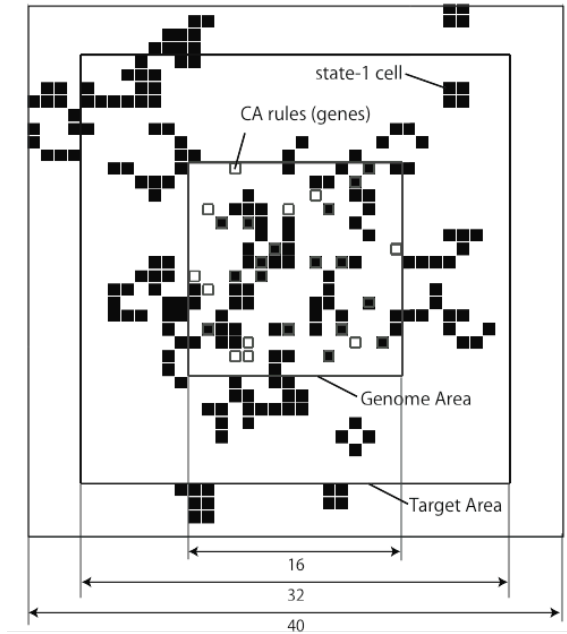

Figure 2: The space used for the simulation. Filled squares represent the state-1 cells. Lined squares represent the positions that the CA rules specify. Those CA rules only exist in Genome Area (inner box). Target Area (outer box) depicts the area in which performance is evaluated in the evolutionary process. While Life rule is applied to the whole space over the Target Area, CA rules are applied within the Genome Area at most.

The rule consists of a 26-bit binary string, 18 bits of which represent the totalistic rule. As is usually the case, the 2D CA rule is specified by the pair of numbers Bx/Sy, which specify when to change the cell's state to 1 and whether its own state is 0 or 1, respectively. For example, the Game of Life is represented by B3/S23. We use 9 bits to represent these two parameters. The remaining 8 bits encode the spatial position of the site controlled by the second layer (4 bits each for x and y coordinate). Each rule specifies a particular single site in the 16×16 cell space. At each time step $t$ of the simulation, first, the Life rule is applied to change the stages from St-1 to St'; then the CA rules on the other layer are applied to override the Life states from St' to St (thus St' does not appear in evaluating fitness for genetic algorithm). The order of the CA rules to be applied is fixed throughout the evolutionary process. When there are multiple rulesets specifying the same site, the later rule always overrides the previous. The total cell space is given as a square of the size 40×40 and the intermediate area controlled by these genes is given as a square of the size 16×16. The target area is defined as a square space and all three squares share a common center. The size of the target area is 32×32 bits and includes the intermediate area (see Fig. 2). The boundary of the space is always set to state-0. A *target area* is also designated, which is significant for the central tasks of the simulation. We will illustrate the detail of the target area and tasks in the next subsection.

## The Target Behavior of the Model

CA rules to perform a particular computational task has been studied previously (Mitchell et al., 1994). Here, we will study homeostatic behaviors observed in the Game of Life, while evolving the ruleset of the intermediate cell space. Instead of a temperature value as in Daisy World, we use the density of cells in state-1 as the target variable to keep constant.

In Daisy World, the system consists of two types of flowers, black and white daisies, with their local temperature values. The black daisies increase the temperature and the white daisies decrease the temper- ature. By setting the growth rate of these flowers according to the local temperature, both flowers show positive feedback effects in their relationship to changing local temperatures. While the black daisies increase simultaneously the temperature and their population, the white daisies simultaneously decrease the temper- ature and increase their population. Loosely linked by the two local temperatures, the global temperature is sustained constantly while both populations of daisies change according to environmental temperature changes. This result shows that the homeostatic behavior is not observed in the state of insensitivity to the environmental stimulus, but is actively achieved by an adaptive coupling of components that are sensitive to the environment.

In order to observe the underlying dynamics of homeostasis in the Game of Life, we constructed two different tasks.

**Reactive task** This controls the density of the target area, making it reactive to the given initial Life pattern density.

**Homeostasis** This sustains the density of the target area regardless of the initial Life pattern density.

The first task is intended to facilitate the development of sensitivity to the given conditions. We tried two conditions for this reactive task: to keep the density of state-1 proportional to the initial density (task $R_p$) and to keep the density inversely proportional to the initial density (task $R_i$). The second task is designed to directly aid the development of homeostatic behavior in the Game of Life patterns. In this homeostatic task, we tested two conditions: maintaining a higher density (task $H_h$) and maintaining a lower density (task $H_l$).

We will see how these behaviors are achieved in the Game of Life space using the evolved CA rules.

| Task | Target density |
|---|---|
| Reactive (proportional) $R_p$ | $d$ |
| Reactive (inversely population) $R_i$ | $1-d_I$ |
| Homeostasis (higher) $H_h$ | 0. |
| Homeostasis (lower) $H_l$ | 0.0 |

Table 1: Notations and numerical values for the target fitness function.

| Parameter | Values |
|---|---|
| number of rules in a genome | 45 |
| population | 30 |
| mutation rate | 0.05 |
| crossover rate | 0.01 |
| mutation rate for genome | 0.01 |
| number of elites | 5 |
| initial density (higher) | 0.5 |
| initial density (lower) | 0.0 |
| evaluated duration (time steps) | 500 |

Table 2: The parameters used in our simulation

## Genetic Algorithm

In order to observe the behaviors, the CA rules encoded in the genome were evolved using a simple genetic algorithm (GA). We prepared 30 genomes in a population, each of which consisted of 45 CA rules initially, that specified each spatial location and the rule content of the 45 totalistic rules in the intermediate area.

For the selection algorithm, first, 5 elites are chosen; then, others are chosen by a roulette selection procedure until the total number of genes reaches 15. Then, those 15 genes are duplicated with mutation and crossover, which modify the spatial locations and the rule content. Mutation occurs at every bit on a genome at probability 0.05 per bit; crossover occurs at probability 0.1 per genome. In addition, the number of genes may also increase or decrease in every generation. A gene is copied or deleted at probability 0.01 per genome. So the number of CA rules (45 in every genome at the beginning of the GA) changes in the evolutionary process.

In the evolving process, the genomes are evolved as a unit against two different initial states with lower and higher density patterns. Here, the lower density is set to 0 and the higher density is set to 0.5. The fitness of the genome is calculated by determining how the average density of state-1 within the target area compares to the specified *target densities*. The target densities in the four tasks are set as follows. Here, $d_I$ is the initial density of state-1 cells, set as either 0 or 0.5. The average of the density is taken from timestep 250 to 500 and summed in both lower and higher density initial condition in the evaluating phase.

Note that we only use a fixed random initial pattern for all evolutionary processes. Once the system evolves, it develops a sufficient generalization capability; the system can do well with new initial patterns. However, a full generalization capability is difficult to obtain. We will revisit this point later. Table 2 shows the parameter values used in this experiment.

## Result

### Evolved Dynamics

In each of four tasks, the genomes in our population were evolved for higher fitness. Figure 3 shows the temporal changes of the state-1 density of the fittest genome in each task. Two lines on the graphs are shown, one for the low initial density case (null pattern) and another for the high density case (0.5). These density values were used in the GA dynamics. The evolved CA rules have characteristic behaviors according to the tasks.

In task $R_p$, the evolved rules change their behaviors in the low and the high densities. The initial low density state almost always creates a sparse pattern in the target area. Using the higher initial density state, the average resultant density state tends to fluctuate around a value of 0.2. Figure 4 shows snapshots of the Life patterns in the two initial densities with the CA rules in task $R_p$. The evolved rules generate cloud-like state-1 Life patterns if the target area is filled with state-1 Life cells, which generally increase the density. On the other hand, from the lower density, the CA rules do not make cloud patterns. The change in the behavior in different initial densities is, therefore, caused by this selective generator of state-1 cells.

In task $R_i$, the densities are altered to be inversely proportional to the initial states. In contrast to the genome of task $R_p$, the evolved CA creates a high-density state when the initial state has a low density, and the density is decreased until all Life patterns are diminished when presented with the high-density initial state. The genome in task $R_i$ behaves like an activator in the initial low-density state, but has to inhibit the spontaneous generator to decrease the state-1 density in the initial high-density state. Figure 5 shows snapshots of Life dynamics with the CA rules in task $R_i$. We can observe that in the low-density case, state-1 cells are spontaneously generated and cloud patterns spread out in the external space of the genome area. In the higher density case, the initial clouded state-1 cells decrease gradually and the Life patterns begin to exhibit static behavior with a small number of fixed patterns. The inversely proportional behavior is achieved by this selective functioning of inhibiting and activating the state-1 cell clouds.

In task $H_h$, the average density cannot reach the target density of 0.5, but the density always reaches the same value of approximately 0.2, regardless of the initial density. We can see that this attracting state is also maintained by the generators of the state-1 Life pattern, seen in the behaviors of tasks $R_p$ and $R_i$. Figure 6 shows snapshots illustrating the CA rules in task $H_h$. Both initial conditions exhibit a similar behavior, with a number of state-1 cells forming cloud-like patterns. This indicates that the evolved CA rules may generate state-1 cells regardless of the initial states.

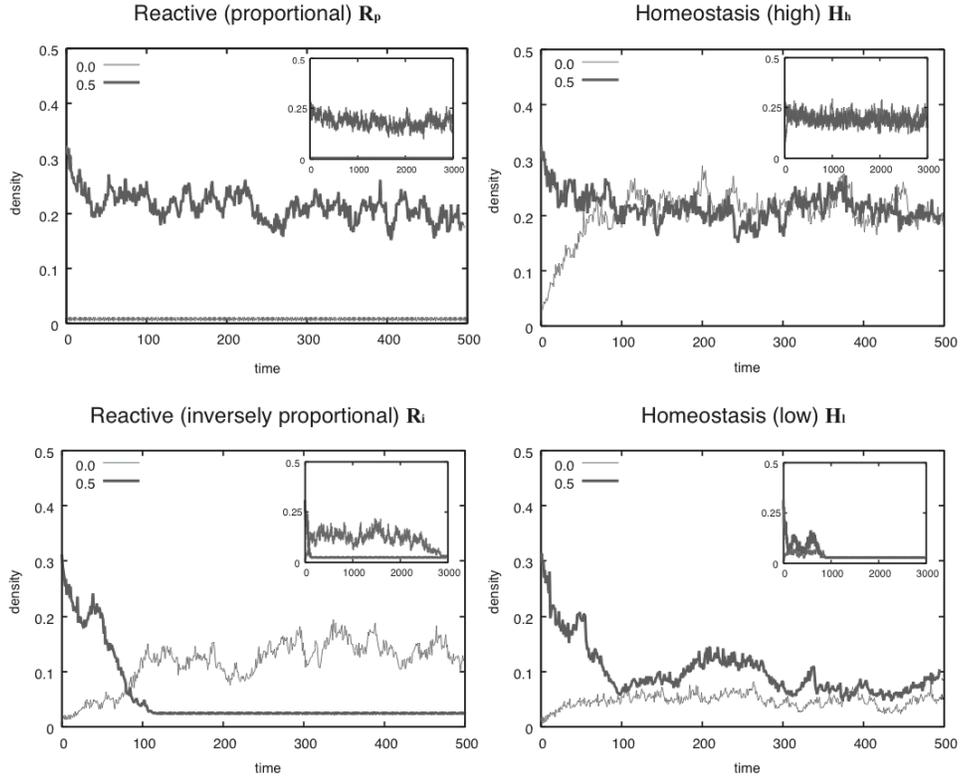

Figure 3: Temporal changes of the state-1 density of the fittest genome in each task. The initial Life patterns used here are the same as those used during the GA procedure. With the higher-density initial state, the state-1 density is kept around 0.2, but with the lower-density initial state, the state-1 density decreases to 0 (task $R_p$). The higher-density initial state shows a state-1 density that drops almost to 0, but the lower initial state causes a growth in state-1 density until the density reaches approximately 0.15 (task $R_i$). When beginning with both low- and high-density initial states, the state-1 densities are maintained at around 0.2 (task $H_h$). When beginning with both initial densities, the state-1 densities are maintained at around 0.05-0.015 (task $H_l$).

In task $H_l$, the densities in both initial conditions approach around 0.05 to 0.10. Figure shows snapshots of dynamics with task $H_l$. The CA rules generate the state-1 cells like the other three rules do. Here, the clouds of state-1 cells are smaller and do not spread out over the whole space, so that density is kept at a low value, but does not reach zero. In both homeostatic tasks, $H_h$ and $H_l$, the evolved rules do not approach the exact target densities. However, they can keep it in tight ranges close to the target densities.

In every case, we can observe CA rules that generate a cloud-like Life pattern to sustain a certain density. Running simulations for longer duration over 500 time steps, we can see two

## Reactive (proportional) R$_p$

initial density : 0.0

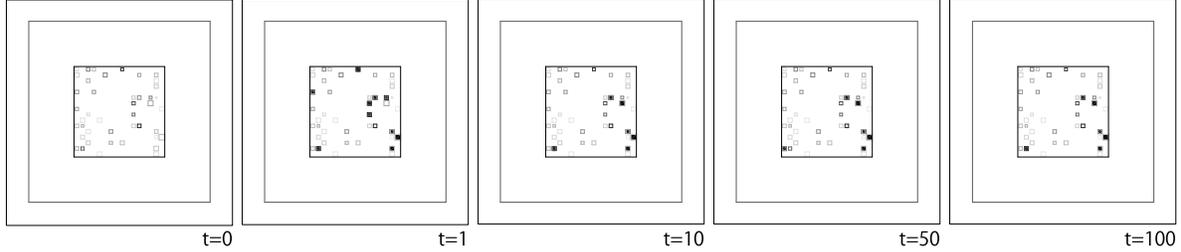

initial density : 0.5

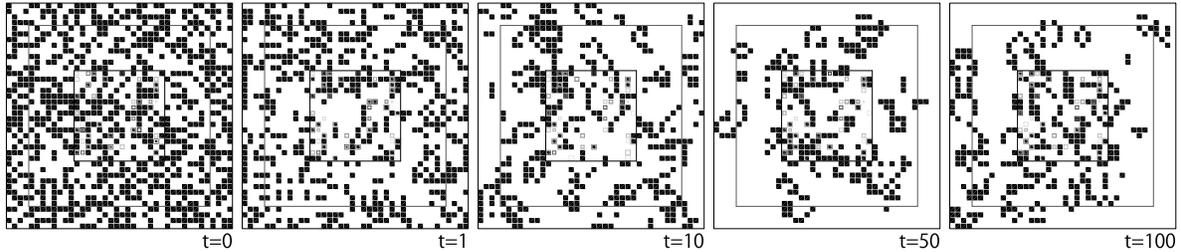

Figure 4: Snapshots of the Life dynamics of the fittest genome for task $H_h$. The top row shows the results when the initial density is zero, whereas in the bottom row the initial density is 0.5. In the zero-density condition, the Life pattern stays almost at zero activity. However, in the higher-density condition, a clouded pattern of state-1 cells is maintained.

different behaviors in these generators. In the tasks $R_p$ and $H_h$, the generator seems to be working infinitely. On the other hand, in tasks $R_i$ and $H_l$, the generator stops working after a finite number of time steps. In the latter two cases, the evolved CA rules have capability to suppress state-1 Life patterns when surrounded by large number of state-1 cells, in particular at the beginning of a run. Such a suppressing behavior would also appear even in the middle of the run if the configurations of Life patterns were accidentally close to the initial configuration. Because of this indiscriminate suppressing capability, in the latter tasks the generators stop after a finite number of time steps.

## Evolved CA rules

Each of the 45 sites in the second layer has a different CA ruleset. One way to characterize these rules is to compute the number of state-1 outputs as a fraction of overall outputs. We define the output bias in the following equation.

$$B_o = N_{state-1}/N_{all} \qquad (1)$$

Here, $N_{all}$ represents the number of all the input conditions, 16 in this case. $N_{state-1}$ is the number of input conditions for making state-1 outputs. The output bias, $B_o$, represents a tendency to create state-1 cells (e.g., the Game of Life has 3/16). We also defined an input bias to see how the number of neighboring state-1 site cells affects the output of state-1 cells. This indicator corresponds to $\lambda$ parameter proposed by Langton (1990).

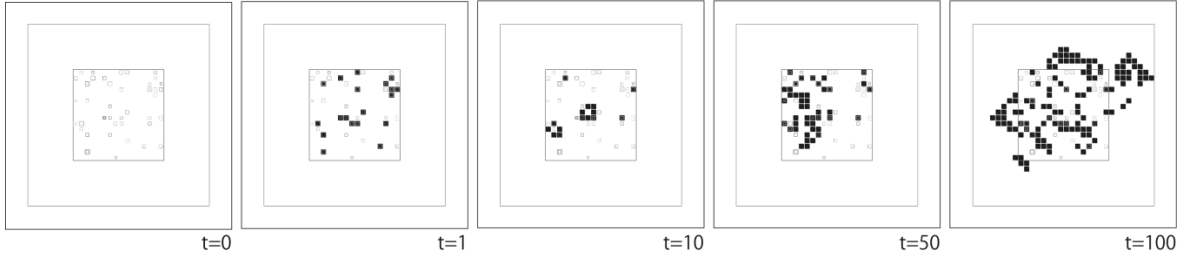
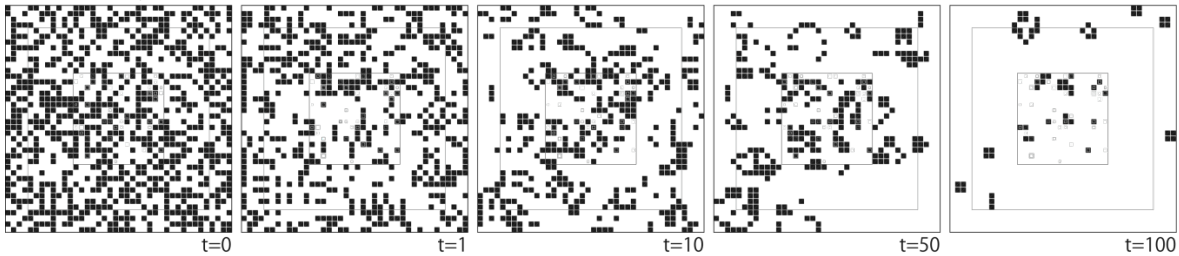

Figure 5: Snapshots of the Life dynamics of the fittest genome in task $R_i$. The top row shows the results when the initial density is zero, whereas in the bottom row the initial density is 0.5. In the zero-density condition, a clouded Life pattern emerges from the initial blank space, but in the 0.5 condition, the clouded pattern of initial state-1 cells eventually disappears.

$$🯄🯄 = 1🯄🯄🯄🯄🯄🯄🯄🯄🯄🯄 – 1🯄🯄🯄🯄🯄🯄🯄 – 1🯄🯄$$

(2)

Here, $C_i$ is the number of the state-1 cells in the neighbors required for state-1 output in each condition. The input bias, $B_i$, represents whether the condition of the number of neighboring state-1 is smaller or larger to create state-1 output; if the value is close to 1, a larger number of neighboring state-1 cells tend to create the state-1, and vice versa (e.g., the Game of Life has 1/3 ). In both indicators, the randomly created genome has normal distribution in 0.5 as a center.

The right parts of each task in Fig. 8 show a histogram comparison of the output bias, $B_o$, and the input bias $B_i$ for the evolved CA rulesets. The left parts of Fig. 8 show the spacial distribution of each CA ruleset. Each square represents the site modified by the CA rule to modify. The size of the square is proportional to the output bias, $B_o$, and the hue of the

## Homeostasis (high) $H^h$

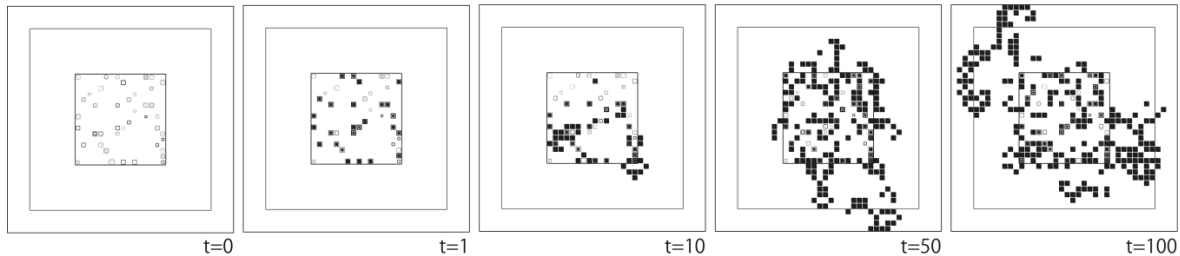

initial density : 0.0

t=0   t=1   t=10   t=50   t=100

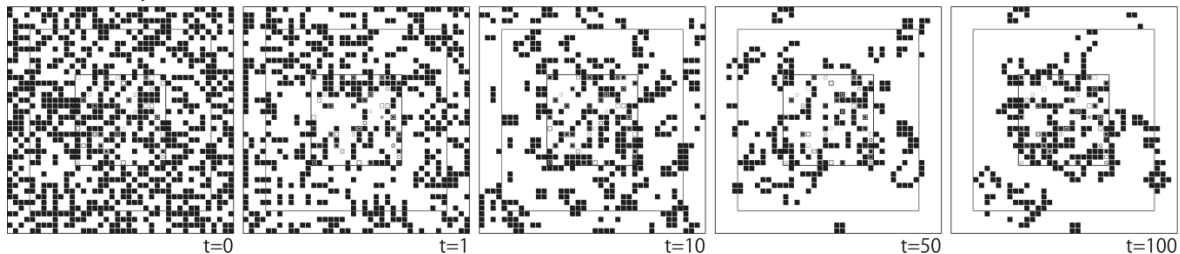

initial density : 0.5

t=0   t=1   t=10   t=50   t=100

Figure 6: Snapshots of the Life dynamics of the fittest genome in task $H_h$. The top row shows the results when the initial density is zero, whereas in the bottom row the initial density is 0.5. In both cases, clouded Life patterns are emerging and spreading out from the genome area to the entire space.

square line represents the input bias, $B_i$.

In tasks $R_p$ and $R_i$, the two indicators have similar distributions for the input biases. They have peaks at relatively smaller values in the input bias histograms. The output bias of the two reactive tasks, $R_p$ and $R_i$ differ to some extent. The output bias is distributed in larger values with task $R_p$ and smaller values with task $R_i$. This indicates that the genome evolved for $R_p$ has a tendency to increase the state-1 cells, but the genome evolved for $R_i$ has the opposite tendency (to decrease the state-1 cells).

In task $H_h$, the output bias is distributed in the higher values and the input is sharply biased in the lower values. This indicates that each rule tends to create state-1 cells from a small number of neighboring inputs. The strong generation of the cloud pattern in task $H_h$ can be explained by this tendency of evolved CA rulesets. In the spatial pattern, the CA rules seem to be distributed uniformly with less variety in terms of types of rules.

In task $H_l$, the rulesets biased to the smaller value in the output bias indicator. The smaller biased output indicates the existence of an inhibitor that suppresses state-1 density. The input bias seems to be distributed at the center, but much more broadly than in other three cases. This means that the CA rules have the ability to respond to a wide variety of local configurations of the state-1 cells, from sparse to condensed conditions. As shown clearly in the diagram of the spatial configuration of each CA rule, the variety of the CA rules in relation to both the input bias and the output bias seems to be much higher in the other three tasks.

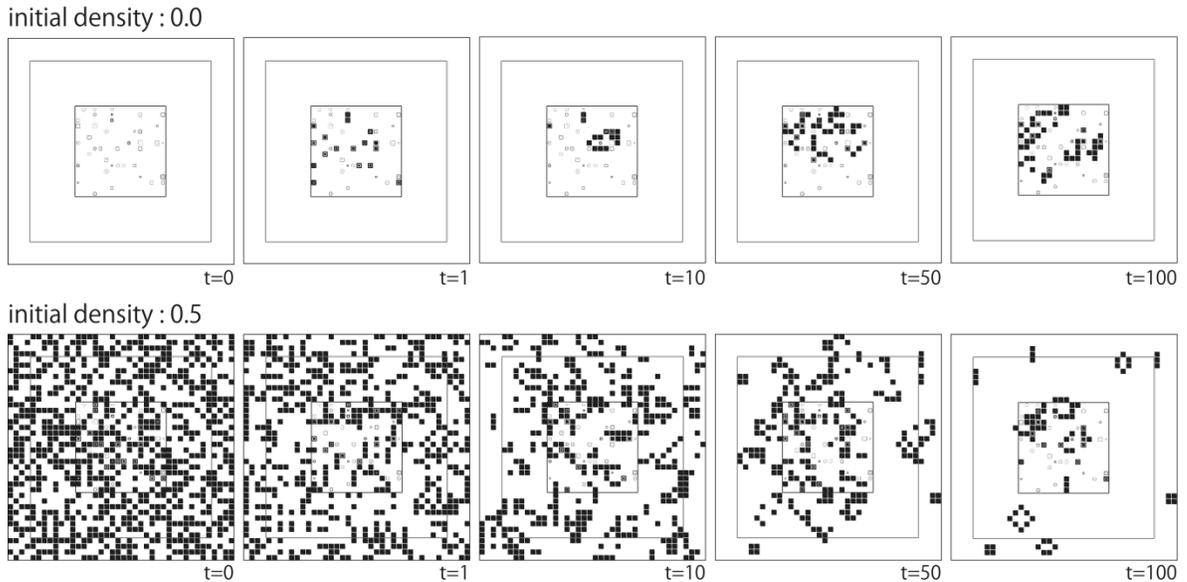

Figure 7: Snapshots of the Life dynamics of the fittest genome in task $H_l$. The top row shows the results when the initial density is zero, whereas in the bottom row the initial density is 0.5. In both cases, the clouded Life patterns seem to be suppressed, but a small group of state-1 cells have emerged in the top of the genome area to fill the entire space. This small cloud pattern does not spread out through the space and stays at the local area, but continues to survive for period of time.

## Generalization

While the genomes under discussion here have been trained with only two fixed initial Life patterns, the final genome acquired some generalization capability. After training the genome set with these two different initial densities of 0.0 and 0.5, we tested the evolved genome against other initial densities. Figure 9 shows the average density after 500 GA time steps for the given initial density. Each density is averaged over 100 different initial Life patterns of a given state-1 density. For a comparison, we also show the average density obtained with only the original Life dynamics, without any second layer.

We can see the generalized abilities with the genomes of all the tasks in the smaller initial densities (< 0.1). Both conditions of the reactive tasks, $R_p$ and $R_i$, have a tendency to increase/decrease the density according to the initial Life density. In task $R_p$, the average density is lineally proportional to the initial density when the initial density is under 0.1. Furthermore, inverse proportion to the initial density can be seen in task $R_i$. With larger initial densities, the averaged densities inevitably increase to a certain density to be dominated by the original Life dynamics. Suppression of the state-1 cells seems to be more difficult than the amplification of state-1 cells.

The genome evolved through task $H_h$ achieves almost identical values of high average density against a wide range of initial densities from 0.0 to 1.0. Compared with the original Game of Life, this evolved genome sustains a higher density, particularly around the lowest and highest initial state. This is achieved by a state-1 generator that increases the state-1 density regardless of the initial states. However, similar behavior can be seen when one adds noise to the Game of Life. Thus the genome does not regulate state-1 density, but rather works as a "random generator". Task $H_l$ has the best suppression ability of the four genomes. Even in relatively higher initial densities, the average density stays under 0.1. In the lower density, the density can be sustained close to the target value of 0.05.

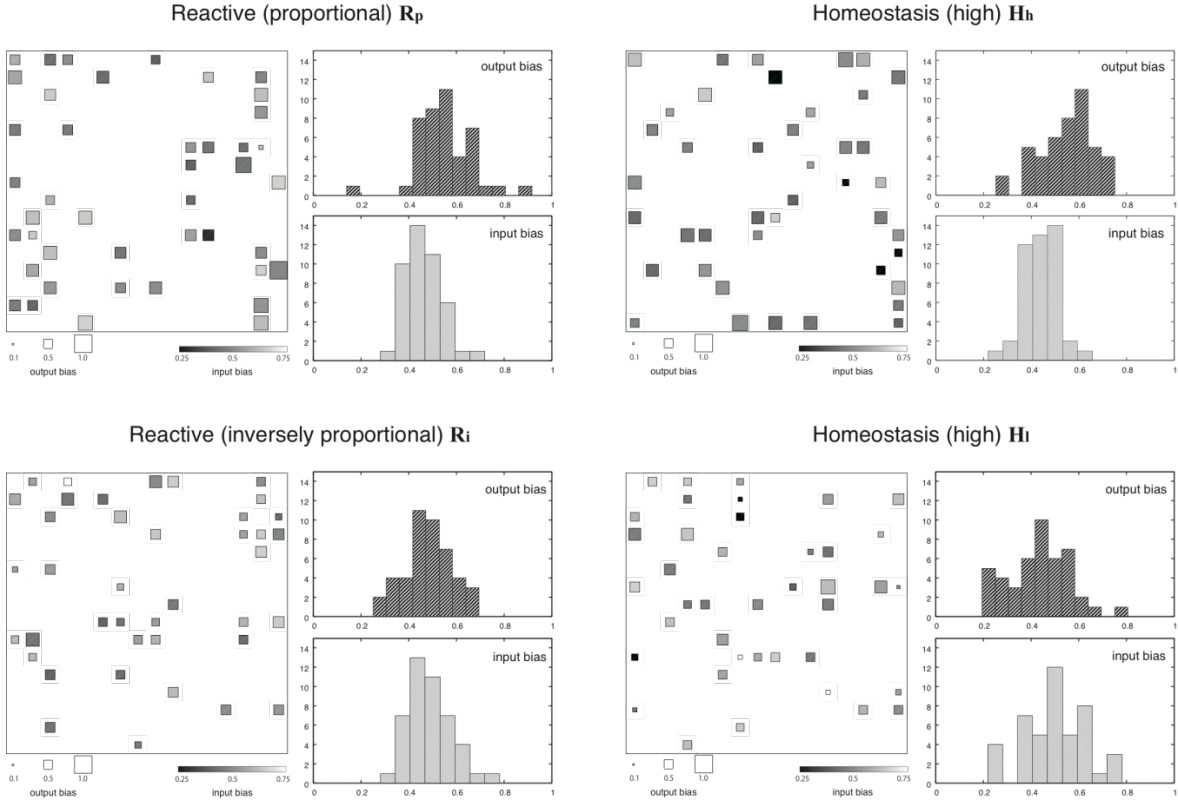

Figure 8: Spatial configurations of the fittest CA rules for each task (left). The size and hue of each square correspond to the input bias (the distribution of the state-1 inputs) and output bias (the distribution of the state-1 outputs), which represent characteristics of the totalistic rules. The larger the output bias, the more frequently the state will become state-1. The larger the input bias, the more state-1 cells are required for its neighboring conditions to become state-1. See the text for details. Histograms of the output bias and the input bias are shown (right).

We are not certain how many attractors this two-layered system has when starting from identical initial density states. Figure 10 shows the distributions of averaged densities for 500 time steps represented as hues. The genome evolved via task $H_h$ has a sharp distribution for density only one attractor exists in the system. Since the CA rules always increase the density, this attractor determines the density's upper boundary in the target area.

There are two different attractors in $R_p$, $R_i$, and $H_l$; One of them is higher density, while the other is lower, almost zero density. In $R_p$ and $R_i$, two attractors are switched by changing the initial density; thus, the proportion/inverse proportion to the initial density can be achieved. On the other hand, the genome evolved in task $H_l$ also has two attractors similar to the genome of $R_i$. The CA rules with $H_l$ work in two different ways, by activating or inhibiting the state-1 cells according to the initial density. This result can be compared to the Daisy World dynamics, in which two components regulate the temperature adaptively.

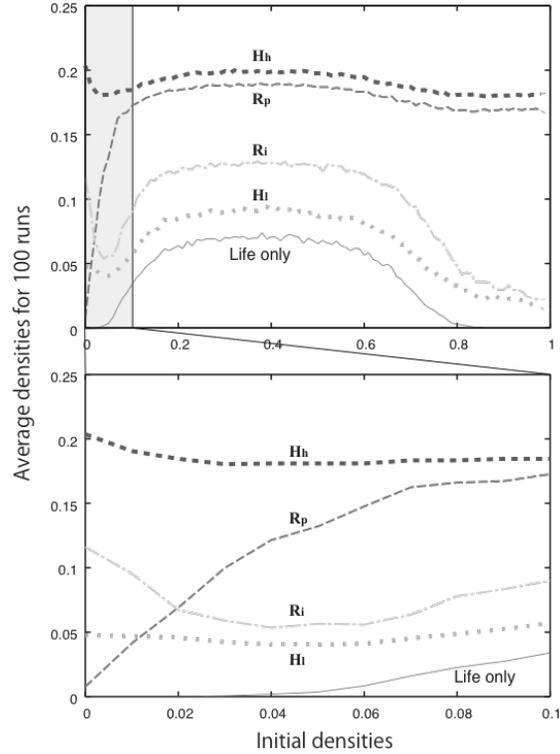

Figure 9: The average densities of 100 samples with different initial Life configurations. The density is calculated over 500 time steps. The evolved genomes in tasks $R_p$, $R_i$, $H_h$, $H_l$ and runs in which only Game of Life rules exist are compared.

Further analysis will be done in the next section to investigate how these CA rules regulate the density, by coupling with the Life dynamics and using spatial cooperation between rules.

## Analysis

In the previous section, the evolved CA rulesets were analyzed in terms of their input/output characteristics. However, the actual functions of these CA rules are tightly coupled with the Life dynamics. In each time step, the CA rule changes the Life state. The life state then specifies a condition for the CA rule in the next time step. This looping process between the evolved CA rules and Life dynamics should be investigated for a full understanding of the evolved CA rules. Moreover, the CA rules do not work alone; rather, spatially adjacent rules collaborate with each other in order to solve the task. We also discuss this spatial collaboration of CA rules in this section.

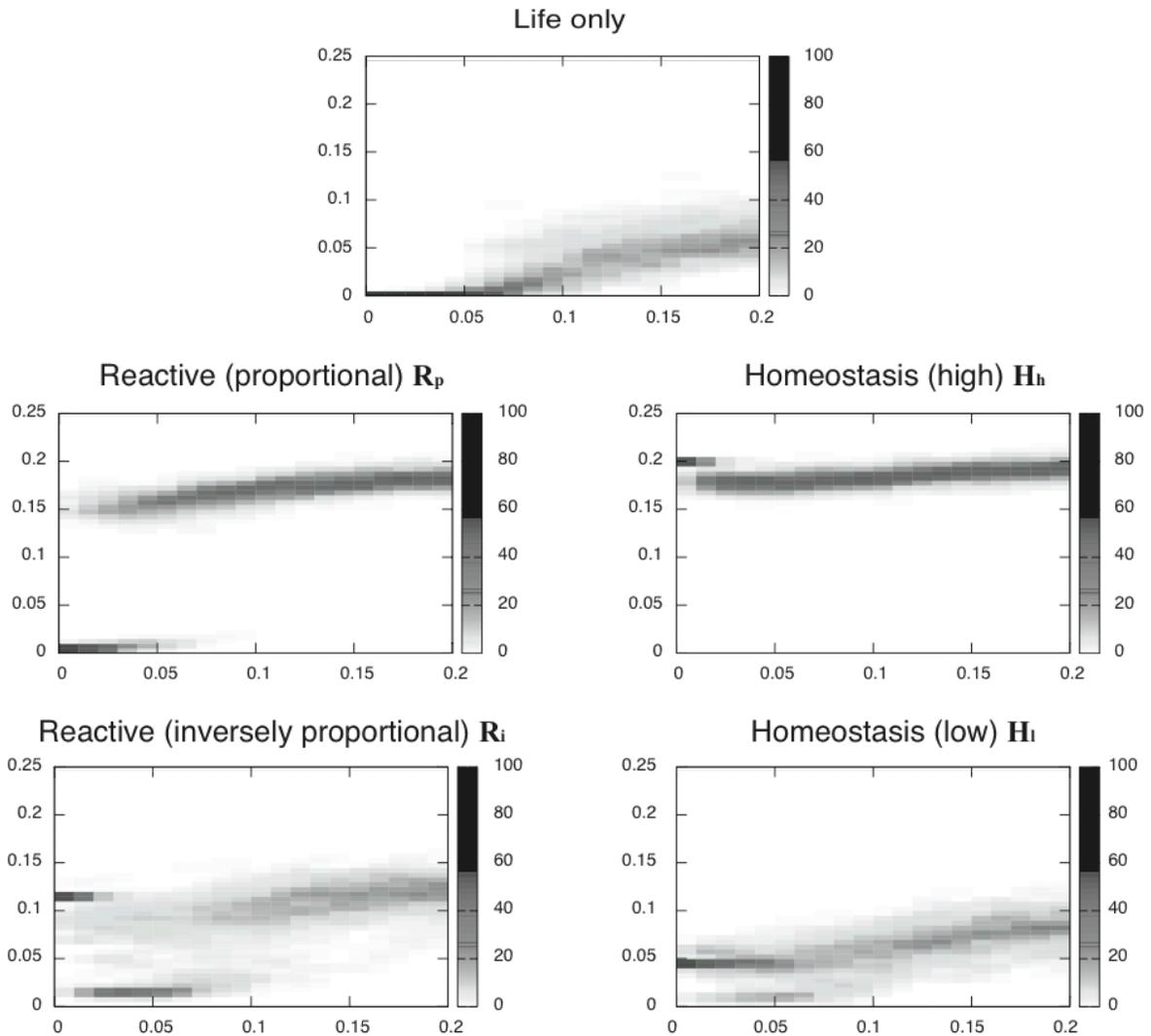

Figure 10: Distribution of average density for 500 steps in the initial density of state-1 Life. One hundred runs were used with a different initial configuration for each density. The hue represents the rate in terms of how many runs go to that density. Except for the case of task $H_h$, we can observe multiple attractors for each density.

## Functions of CA rules

As seen in section 3.2, the evolved CA rules have a variety of input/output characteristics. The input/output indicators can to some extent show how these rules work. However, the actual functions of these evolved rules have not yet been separated from their ongoing process in coupling with Life dynamics. Therefore, we first investigated significance of each CA rule for regulating the state-1 density by disabling each CA rule and observing the effect. Figure 11 shows relative changes of the average density in the target area when each CA rule is removed. The hued squares represent the specific sites of the CA rules. The hue indicates how much the average density shifts, increases, or decreases when removing the rule, compared with the original density. Thus, if the value is high, the CA rule has the function of decreasing the density by suppressing state-1 cells. Conversely, if the value is low, the CA rule has the function of increasing the density by promoting state-1 cells. In each task, we investigate the same initial configuration applied to the evolutionary process; the densities are 0.0 and 0.5.

We can see clear differences in the four tasks, illustrating how each CA rule contributes to regulating the density. The CA rules evolved in task $R_p$ do almost nothing with the lower initial condition (top left in Fig.11). Most rules do not affect the density by removing it. With the higher initial condition, all rules have minus changes, shown by the darker hue. This means these CA rules are all activating state-1 cells so that the density is increased. Conversely, in the case of task $R_i$, shown at the bottom left of Fig. 11, it is clear that almost all the activating rules in the lower initial condition (shown in green or blue) work as the inhibitors in the higher initial condition.

These two results correspond exactly to the behavior obtained in the reacting tasks. The activation and inhibition (or idle) functions of the state-1 patterns are chosen according to the initial densities. These differences in function in the two conditions are not explained by the static analysis of the CA rules, but are explained as a coupling of rules and Life in spatial and temporal dynamics. The genome evolved in task $H_h$ has many rules with minus values, indicating an activating function, in both initial conditions (top right in Fig.11). The function of every rule for activating state-1 cells seems to be identical in the two different conditions, unlike in the above two reactive tasks, $R_p$ and $R_i$. This corresponds to the result of the genome analysis in the previous section, where the genome for task $H_h$ has many state-1 output CA rules.

On the other hand, the CA rules with task $H_l$ have a more complex composition. Interestingly, the activating and inhibiting functions coexist in one genome. It is observed that strong activators exist in the center of the space, and inhibitors exist in the peripheral areas of the left and bottom. We assume that

state-1 cells are created by the activators, but at the same time, the surplus state-1 cells are reduced by the inhibitor so that the average density is kept at a constant value. In the higher density case, this periphery inhibitor works as a barrier to protect the state-1 generator from noisy patterns coming from outside the genome area.

The analysis indicates that there are two types of regulating mechanisms used by evolved CA rules in the Game of Life space. The rules evolved in the reactive tasks, $R_p$ and $R_i$, have mostly an homogeneous property (increase or decrease) for each initial density. However, their activities drastically change according to changes in the initial patterns. On the other hand, the CA rules in the $H_l$ have identical activity in the different initial conditions because both activating rules and inhibiting rules co-exist in the same space. The strategy taken here is explained by regulation being achieved by preparing a variety of CA rules that work as activators and inhibitors. The rules evolved in task $H_h$ do not actually regulate the density. They have the homogeneous function of activator in both initial conditions.

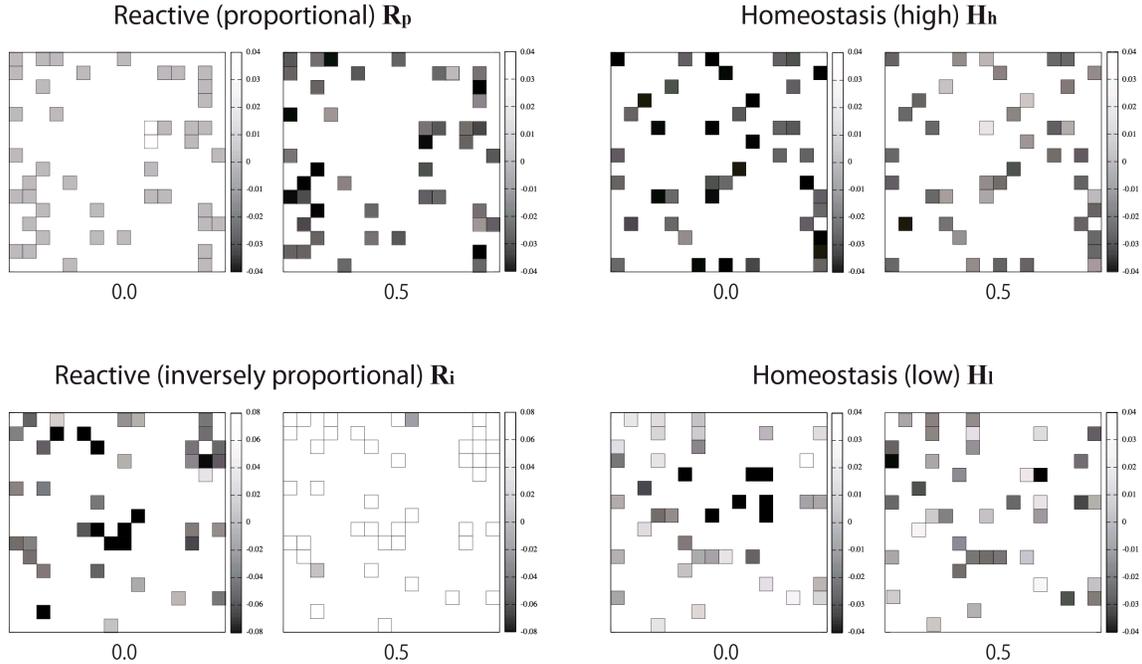

Figure 11: Functions of the CA rules evolved in task $R_p$ are shown in hued squares. The hue of the square represents the "effective function" for regulating entire density. Namely, by calculating the change of the total average density by removing the CA rule of each site, we notice that each rule has apparent functions in reactive conditions, but in case of homeostasis of the lower density case, positive and negative tendencies coexist.

## Spatial Regulation

The evolving CA rules use the totalistic rules to update the Life site to state-1/state-0. The function of the evolved rules is not only determined by these explicit rules, but also by their spatial configuration. To investigate this implicit spatial effect, we tried to test the behaviors of evolved CA rules with the same totalistic rules, but with randomized positions. Figure 12 shows the histogram of the density in relation to the initial density when the CA rule's positions are randomized for each trial. The solid line shows the average density of the histogram in the randomized-position case, while the dotted line shows that in the evolved case.

In task $H_h$, the average density has almost the same distribution as that of the non-randomized cases shown in Fig. 10. This means that spatial configuration does not have a significant role in sustaining the high density, but each activating function of the CA rule contributes toward the increase in density.

On the other hand, in the reactive tasks, $R_p$ and $R_i$, the distributions are changed entirely from the non-randomized conditions. The sensitivities to the initial density are lost by randomizing the positions of the CA rules. The two attractors, as seen in the revolved case, seem to exist in the randomized position, but the ways in which the initial density is sorted into these attractors are totally changed. This indicates that the sorting ability of the two

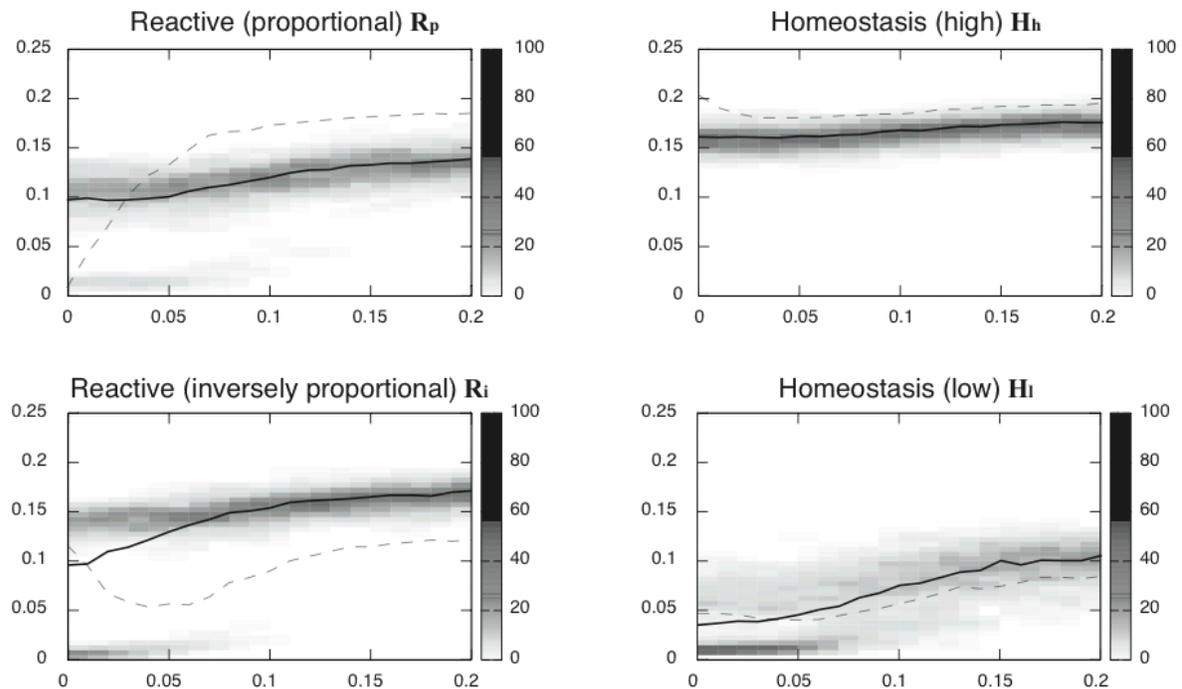

Figure 12: Histogram of the densities calculated from the 100 samples with the CA rules, in which positions are randomized. This figure should be compared with Fig.10. The solid lines show the averaged density calculated from all 100 samples in the randomized case, while the dotted lines show the same in the non-randomized case. Distribution becomes much broader compared with the evolved cases.

attractors is achieved by the configurations of the CA rules.

In the case of task $H_l$, the averaged density in the randomized case (shown in a solid line), has not changed very much compared with the evolved case (shown by the dotted line). On the other hand, the histogram shows that the distribution of the randomized case becomes much broader than that of the evolved case. We assume that a variety of CA rules evolved in the case of homeostasis $H_l$ condition, the homeostatic property is maintained even in the randomized position experiment.

### Evolving CA Rules in Fixed Positions

In the previous subsection, we showed that the regulation of the average density relies not only on the explicitly coded totalistic rule, but also on spatial configurations of the CA rules. Here, to investigate this spatial effect in the evolutionary process, the CA rules for solving the same tasks are evolved using different settings. In them, the CA rules are aligned in fixed positions for the entire evolutionary process. We used two different configurations of fixed CA rules: filling the whole genome site with 256 CA rules and filling a checker configuration with 128 CA rules.

Figure 13 shows the fitness changes in the two fixed-space simulations and the original

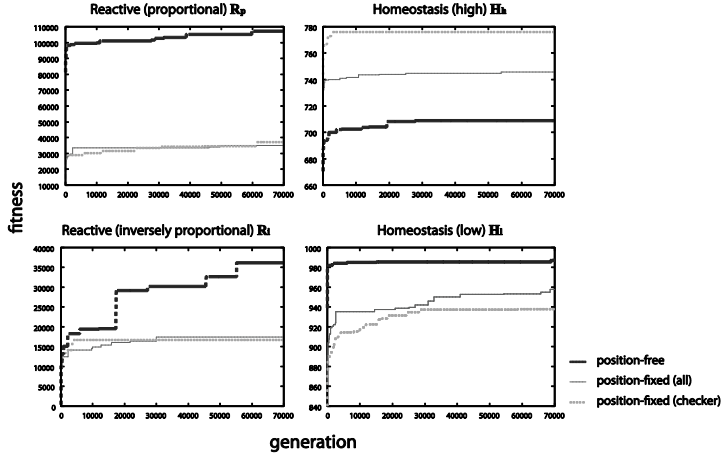

Figure 13: The fitness changes in four tasks for position-free and position-fixed settings. In the position-fixed settings, only the totalistic rules are evolved, but the positions of the rules are fixed. For the position-fixed settings, there are two configurations of CA rules; all positions and a checker pattern. In task $R_p$ and $R_i$, the fitness in the position-fiexed case never increases more than in the position-free cases. In task $H_h$, the fitness in the position-fiexed case is constantly higher than in the position-free settings. In $H_l$, with the checker pattern setting, the fitness in the position-fiexed case increases to approximate the fitness of the position-free case.

position-free simulations. The result shows that in the two reactive tasks ($R_p$, $R_i$), the fitness increases much less than in the position-free settings. In the homeostasis task, however, the fitness increases much more than in the original one. Why is it that the CA rules with tasks $R_p$ and $R_i$ cannot evolve in the position-fixed setting? There might be some difficulties for these tasks in the position-fixed setting. First, randomly generated CA rules have a tendency to create a larger number of state-1 cells compared to the Life rule. With this property, it is difficult to sustain the lower density of the state-1 state in the genome space. As a result, it is not possible to obtain better fitness with tasks $R_p$ and $R_i$ because the CA rules cannot decrease the density in any case. Second, the reactive tasks require sensitivity to the initial configuration of Life. To have this sensitivity, the CA rules have to "see" the Life dynamics by coupling their own dynamics with Life dynamics. In the case of position-free settings, the CA rules are relatively sparse and there is some room for the Life to behave according to its own dynamics. However, in the case of position-fixed setting, there is no room

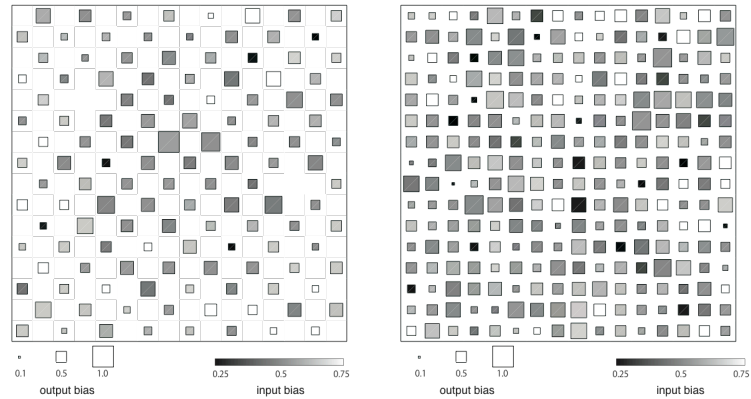

Figure 14: CA rules evolved in position-fixed settings for task $H_l$ are shown. In both patterns, state-1 cells are bred only in the middle of the space, but suppressed in the surrounding area.

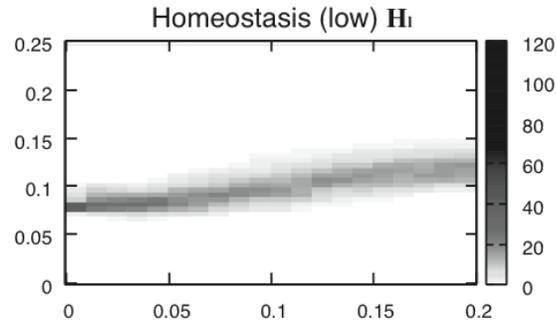

Figure 15 depicts the generalization ability of the evolved genome in the position-fixed setting by showing the average density in the genome. It shows that the generalization ability is much higher with the position- fixed setting than with the position-free setting. This may be because plenty of inhibitory rules in the surrounding area can perfectly cancel out the external noise patterns, so that the internal activators can easily achieve a constant density.

for the Life dynamics — only the evolved CA rules govern the genome area. Consequently, the relationship between the CA rules and Life is one directional.

The CA rules evolved in task $H_l$ are illustrated in Fig. 14. In both the checker pattern and the filled pattern, the CA rules with $H_l$ have the larger state-1 output rules in the middle of the space — these larger output rules are surrounded by smaller output rules (depicted by smaller squares). This behavior seems to be the same strategy seen in the genome in the position-free setting, but it shows more clearly, showing that the activators at the central position are surrounded by the remaining inhibitors.

Figure 15 depicts the generalization ability of the evolved genome in the position-fixed setting by showing the average density in the genome. It shows that the generalization ability is much higher with the position- fixed setting than with the position-free setting. This may be because plenty of inhibitory rules in the surrounding area can perfectly cancel out the external noise patterns, so that the internal activators can easily achieve a constant density.

## Discussion

In this paper, we studied homeostasis and adaptability with respect to cell states in the Game of Life, as opposed to temperature as in Daisy World. There are some lessons to be learned here, both about homeostasis and the effect of noise.

First, we evolved reactive tasks that required responses to the initial Life density in relation to two different settings (proportional ($R_i$) and inversely proportional ($R_i$)). In tasks $R_p$ and $R_i$ there have evolved appropriate behaviors in which the CA rules increased or decreased the average density according to the initial density. The strategy was to generate cloud-like state-1 patterns, which tend to increase the average density. They selected whether to generate them or not according to the initial densities. The evolved CA rules also showed generalization for changing the initial patterns to unused configurations in the evolutionary processes.

The CA rules evolved in task $H_h$ also had the capability to generate a state-1 density. However, they generated the state-1 cells regardless of the initial Life density. By generating state-1 cells continuously, a high average density was sustained. The behavior of this rule can be compared to a "random generator" that randomly updates the cell states at a certain probability. In task $H_h$, the evolved CA ruleset mimicked a random generator using a deterministic rule.

The CA rules evolved in task $H_l$ had interesting properties. By examining the generalization ability, it was found that the rules can sustain the average density for the initial density from 0 to 0.1. In contrast to the case of $H_h$, this homeostatic behavior is achieved using various types of CA rules. Some CA rules tend to generate more state-1 Life patterns than others and increase the state-1 density as a result, which we call "activator" rulesets. In contrast, some CA rules show the opposite behavior and lower the state-1 density — we call these "inhibitor" rulesets.

It is noted that the CA rules are not independent of the Life dynamics. As shown in section 4.2, the effective properties of the CA rules are determined by the state-1 spatial patterns around them. So we say that certain spatial patterns generate more state-1 patterns with the activator ruleset, and other patterns absorb state-1 patterns with the inhibitor ruleset. The spatial patterns are, thus, entangled with the CA rules, working as activators or inhibitors.

Those spatial patterns continuously generate themselves within some kind of viability range which is determined by the local CA rulesets. And when the pattern goes beyond this viability zone, then they will die, so that the system just undergoes the original Life rule. This situation can be seen in Fig. 10 as the existence of two attractors in task $H_l$. In the lower initial densities, when the patterns survive, the average densities are kept in a certain range. However they sometimes become extinct, which results in the average densities going to zero under the Life rule.

In this point of view, some spatial pattern could be regarded as the daisies in the Daisy World. We simply take the activating spatial patterns as black daisies and the inhibiting spatial patterns as white daisies. The black and white daisies, in Daisy World, have the opposite responses to the sunlight, they can self-regulate the temperature by tuning their population size. If there are more black daisies, the temperature goes up; if there are more white daisies, the temperature goes down. Here, in our result, the activating patterns make temperature(state-1 density) higher, while inhibiting patterns make it lower. The system regulates population of those patterns according to the sunlight(initial state-1 density). The correspondence between the Daisy World and this Game of Life system is shown in Table 3. rates of daisies are determined by the local temperatures. The concept

| Daisy World | Our model |
|---|---|
| external temperature | initial state-1 density |
| averaged temperature | averaged state-1 density |
| black daisy | state-1 generating spatial pattern |
| white daisy | state-1 absorbing spatial pattern |
| a growth function of daisies | emerging dynamics of spatial patterns |

Table 3: A comparative chart between the Daisy World model and the present CA model

In the Daisy World, the growth of temperature is not implemented in our system explicitly. Instead, local Life patterns and the CA rulesets determine how spatial patterns grow in the following timesteps. Note that the equivalent of Daisy Worlds local temperature in our system is not just a one-dimensional variable that explicitly specifies the growth rate of state-1, but is instead a spatio-temporal pattern that drives responses from the evolved CA rulesets. This dynamical property of Daisy World has also been discussed in our previous model of the mobile daisy agent (Ikegami and Suzuki, 2008).

We can obtain further implications of this by focusing on the spatial configuration of the evolved CA ruleset. In section 4.2, we examined how the spatial configuration of the evolved CA rules affected the entire function. We found that by randomizing the positions of the CA rules, their functions responding to the initial density is totally lost in the case of the two reactive tasks. This result indicates that not only are the explicit rules coded as a totalistic rule, but their positions also have a significant role in solving the given tasks. In particular, the CA rules for task $H_l$ have a characteristic configuration such that the activating rules are localized at the center, while the inhibiting rules surround the activator. This configuration (Fig. 11) can be explained by the two opposite mechanisms. By definition of homeostasis, the system must generate more state-1 cells with the low density initial state, and it must remove state-1 cells with the higher density initial state. Here, these two behaviors are incompatible because the former requires activating behavior and the latter requires inhibitive behavior. One solution to this contradiction is the spatial differentiation seen in the evolved CA rules.

These results remind us of the discussion of the boundary between organisms and environment. In our model, we defined the genome area as a controllable section in the system. However it does not mean a bound- ary is exactly corresponding to this genome section. When we tried to define an organism in the system, the boundary between the organism and environment changed dynamically in the Game of Life layer; it shrinks and broadens, and sometimes disappears. In our results, the boundary is not clearly seen in tasks $R_p$, $R_i$, and $H_h$, but we can see it in task $H_l$ as a membrane-like function of inhibiting the external noise patterns at the peripheral region, while state-1 clouds are concurrently generated inside it. Varela and Maturana argued that the importance of creating boundary processing through the model's own organization, like a membrane of a cell (Varela, 1979). It is actively discussed that how much significant is in regenerating boundary to an- imate living organism (Bourgine and Stewart, 2004, Moreno and Etxeberria, 2005, Paolo, 2005). To clarify its importance, we need alternative realizations of the membrane system. Our system is a first small step in that direction.

The significance of this study may also lie in observing the homeostatic regulation of discrete pattern dynamics of Life. So far, the evolved CA rulesets use a simple strategy that locally increases or decreases the density of state-1 cells, so that the entire state enters complicated, chaotic patterns. Instead of using such chaos-like attractors as were employed in this study, it may be interesting to use unique Game of Life creatures such as oscillators, breeders, and guns to generate homeostasis in more complex ways. We might then expect to see alternative homeostatic mechanisms that are very different from those seen in Daisy World.

# References

(Ashby, 1960) Ashby, W. (1960). *Design for a Brain: The Origin of Adaptive Behaviour*. Chapman and Hall, London. (Bernard, 1957) Bernard, C. (1957). *An Introduction to the Study of Experimental MedicineThe wisdom of the body*. Dover Publications, New York.

(Bourgine and Stewart, 2004) Bourgine, P. and Stewart, J. (2004). Autopoiesis and cognition. *Artificial Life*, 10:327–345. (Cannon, 1932) Cannon, W. B. (1932). *The wisdom of the body*. W.W. Norton and Company, New York.

(Harvey, 2004) Harvey, I. (2004). Homeostasis and rein control: From daisyworld to active perception. In Pollack, J., Bedau, M., Husbands, P., Ikegami, T., and Watson, R. A., editors, *Artificial Life IX : Proceeding of the 9th International Conference on the Simulation and Synthesis of Living Systems*, pages 309–314. Cambridge, MA: MIT Press.

(Ikegami and Suzuki, 2008) Ikegami, T. and Suzuki, K. (2008). From homeostatic to homeo dynamic self. *BioSystems*, 91(2):388–400.

(Langton, 1990) Langton, C. G. (1990). Computation at the edge of chaos: phase transitions and emergent computation. In *CNLS '89: Proceedings of the ninth annual international conference of the Center for Nonlinear Studies on Self-organizing, Collective, and Cooperative Phenomena in Natural and Artificial Computing Networks on Emergent computation*, pages 12–37, Amsterdam, The Netherlands, The Netherlands. North-Holland Publishing Co.

(Lovelock, 1972) Lovelock, J. E. (1972). Gaia as seen through the atmosphere. *Atmos. Environ.*, 6:579–580.

(Mitchell et al., 1994) Mitchell, M., Crutchfield, J. P., and Hraber, P. T. (1994). Evolving cellular automata to perform computations: Mechanisms and impediments. *Physica D*, 75:361–391.

(Moreno and Etxeberria, 2005) Moreno, A. and Etxeberria, A. (2005). Agency in natural and artificial systems. *Artificial Life*, 11:161–175.

(Paolo, 2005) Paolo, E. D. (2005). Autopoiesis, adaptivity, teleology, agency. *Phenomenology and the Cognitive Sciences*, 4:429–452.

(Taylor, 2004) Taylor, T. (2004). Redrawing the boundary between organism and environment. In Pollack, J., Bedau, M., Husbands, P., Ikegami, T., and Watson, R. A., editors, *Artificial Life IX : Proceeding of the 9th International Conference on the Simulation and Synthesis of Living Systems*, pages 268–273. Cambridge, MA: MIT Press.

(Varela, 1979) Varela, F. R. (1979). *Principles of Biological Autonomy*. New York: North Hollandk.

(Watson and Lovelock, 1983) Watson, A. J. and Lovelock, J. E. (1983). Biological homeostasis of the global environment: the parable of daisyworld. *Tellus*, 35B:284–289.

(Wiener, 1948) Wiener, N. (1948). *Cybernetics or Control and Communication in the Animal and the Machine*. Paris: Hermann et Cie - MIT Press, Cambridge, MA.s